# Building a Question Answering System for the Manufacturing Domain

Liu Xingguang[1], Cheng Zhenbo[1,*], Shen Zhengyuan[1], Zhang Haoxin[2], Meng Hangcheng[2], Xu Xuesong[1], Xiao Gang[1]

[1]College of Computer Science and Technology and College of Software Engineer, ZheJiang University of Technology, HangZhou, ZheJiang, 310032, China.
[2]College of Mechanical Engineering, ZheJiang University of Technology, HangZhou, ZheJiang 310032, China.

∗Corresponding author. Email address: czb@zjut.edu.cn (Cheng Zhenbo)

## Abstract

The design or simulation analysis of special equipment products must follow the national standards, and hence it may be necessary to repeatedly consult the contents of the standards in the design process. However, it is difficult for the traditional question answering system based on keyword retrieval to give accurate answers to technical questions. Therefore, we use natural language processing techniques to design a question answering system for the decision-making process in pressure vessel design. To solve the problem of insufficient training data for the technology question answering system, we propose a method to generate questions according to a declarative sentence from several different dimensions so that multiple question-answer pairs can be obtained from a declarative sentence. In addition, we designed an interactive attention model based on a bidirectional long short-term memory (BiLSTM) network to improve the performance of the similarity comparison of two question sentences. Finally, the performance of the question answering system was tested on public and technical domain datasets.

# 1 Introduction

The term special equipment refers to boilers, pressure vessels (including gas cylinders), pressure pipelines, elevators, hoisting machinery, passenger ropeways, large amusement facilities, and special motor vehicles on sites that involve threats to human safety and high risk. The design process of special equipment products often needs to follow various national standards (/, 2003). Taking China's elevator products as an example, the whole life cycle of an elevator product must comply with the related equipment safety regulations. In addition, elevator products must follow 12 safety technical specifications and 35 national and industrial standards. These regulations, technical specifications, and standards ensure the safety of products with respect to the design, manufacture, installation, and maintenance of elevators. Therefore, many decisions in the design and manufacturing process of these products need to be made by consulting various standards and technical specifications frequently.

The design and manufacture of mechanical products involve many types of decision-making problems, and artificial intelligence technology is often used to assist these decision-making processes to improve the efficiency of decision-making and the accuracy of the decision-making results (D.-H.Kim et al., 2018). Early decision-making systems mainly consisted of parametric systems and expert systems. A parametric system is primarily used to determine the selection of parameters in the design process (A.K.Sood et al., 2010). The parameters can be automatically obtained under certain constraints through an intelligent algorithm. Systems that use intelligent algorithms to assist parameter selection have been successfully applied in aircraft design, automobile design, and elevator design. In an expert system, a process of formalizing human expert knowledge is used to form rules, and then the rules are used to assist decision-making. Expert systems have been successfully applied in the design and manufacture of mechanical products (K.S.Metaxiotis et al., 2002). However, both parametric and expert systems, which are difficult to use for decision-making in certain domains, can only assist decision-making for specific products.

One of the most promising ways to make decisions within the technical domain is through the use of a question answering (QA) system (W.Yu et al., 2020). A QA system aims to provide accurate answers to users' questions in natural language. This task has a long history and dates back to the 1960s (M.A.CalijorneSoares & F.S.Parreiras, 2018). The goal of QA systems is to give the final answer to a question directly instead of returning a list of relevant fragments or hyperlinks, thus providing better user-friendliness and efficiency than search engines (F.Zhu et al., 2021). The technical basis of QA includes natural language processing, information retrieval, and information extraction. The traditional QA system generally includes three processing stages, namely question analysis, document retrieval, and answer extraction. The purpose of question analysis is either to obtain the queries that can be used in the subsequent stage of document retrieval or to classify the question to guide the subsequent stage of answer extraction (F.Zhu et al., 2021). The purpose of document retrieval is to search for documents or paragraphs related to the queries of the question from the text dataset. Answer extraction is used to further filter documents or paragraphs to find those closely related to questions on the basis of the previous stage. Therefore, a traditional QA system only returns the documents related to the answers to the questions, rather than returning the answers to the questions directly.

With the development of deep learning and large-scale pretrained language models, domain-oriented QA systems have made great progress (Z.Huang et al., 2020). Generally speaking, these systems need to construct a question-answer pair dataset in advance. When a new question is given, the answer to the question that is most similar to the new question can be taken as the answer to the new question by comparing the similarity between the new question and each question in the dataset. Therefore, the main factors affecting the performance of QA systems are the size of question-answer pair dataset and the question similarity comparison algorithm. However, question-answer pair datasets in the mechanical manufacturing field tend to be small in scale, and it is difficult for existing sentence similarity comparison algorithms to capture the semantics of sentences associated with the application context. As a result, it is difficult to achieve the expected results using current QA systems in the mechanical manufacturing field.

To address the aforementioned challenges, we propose a QA framework that includes an interactive attention model based on a BiLSTM network to improve the performance of the similarity comparison of two questions. In addition, to solve the problem of insufficient training data in a target domain QA system, we propose a method to generate questions according to a declarative sentence from several semantic dimensions so that multiple question-answer pairs can be accumulated from one declarative sentence. Experiments show our proposed system can provide superior performance than other state-of-the-art methods.

## 2  Related Research

**2.1  Methods of building a QA system**

The aim of a QA system is to provide precise answers in natural language according to the user's question. The initial idea of the QA system began with the Turing test in 1950 (F.Zhu et al., 2021). In the 1960s, structured QA systems were first proposed (F.Zhu et al., 2021). The main process of a structured QA system is to analyze the questions, turn them into database queries, and finally search for the answers in a database. The most well-known system from this period is the Baseball system , which can answer questions related to the American Basketball League in a certain season (B.F.Green et al., 1961). In the 1990s, due to the rapid development of the Internet, users produced a large amount of text data. The knowledge of QA systems comes from various fields, and the QA system in this period gradually entered the open domain (F.Zhu et al., 2021; K.S.D.Ishwari et al., 2019; T.Sultana & S.Badugu, 2020).

With the success of deep learning in many application fields, deep learning has been widely used in all stages of QA systems (Z.Huang et al., 2020). In terms of question analysis, many studies have developed a classifier based on neural networks to classify questions. For example, the classification of a given question can be realized by a convolutional neural network (T.Lei et al., 2018) or LSTM (W.Xia et al., 2018). For document retrieval in QA systems (O.Khattab et al., 2021; V.Karpukhin et al., 2020), to reduce the difficulty of accurately matching terms, problems or documents can be encoded into a vector space to avoid the decrease in retrieval performance caused by mismatching terms. For example, (R.Das et al., 2019; Y.Feldman & R.El-Yaniv, 2019) proposed measuring the similarity of documents or questions by training the encoder to encode documents or questions into vectors and then directly calculating the inner product of the vectors.

The pretrained model of natural language processing has led to remarkable breakthroughs in various natural language tasks (X.Qiu et al., 2020). Pretrained models such as bidirectional encoder representations from transformers (BERT) have also been applied in QA systems (M.Zaib et al., 2020). The question-answer pairs can be constructed in advance, and then the answer to the new question can be obtained by comparing the similarity between the new question and existing questions. This method obtains the answers to questions through an end-to-end training process. Their performance mainly depends on the quality of the question-answer pairs and the precise similarity comparison between questions. BERT can be used to obtain a vector representation of the questions using masked language modeling in the pretraining stage, and then the similarity between the questions can be obtained by fine-tuning the model parameters in the fine-tuning stage (C.Qu et al., 2019; S.Gupta, 2019; W.Sakata et al., 2019).

Recently, the two-stage retriever-reader QA framework proposed by (D.Chen et al., 2017) has attracted extensive attention. Its retriever quickly retrieves several candidate documents related to a given problem from a large number of documents. Its reader then infers the answer to the given question from the candidate document. The aim of research on two-stage QA systems is to improve the performance of these systems using different approaches, such as the development of pre-training methods (K.Guu et al., 2020; K.Lee et al., 2019), semantic alignment between questions and documents (L.Wu et al., 2018; V.Karpukhin et al., 2020), interactive attention based on BERT (M.Gardner et al., 2019; W.Yang et al., 2019), and normalization among multiple documents (Z.Wang et al., 2019).

**2.2 Application of QA systems in manufacturing**

Early studies on QA systems in manufacturing mainly realized QA systems as expert systems. Expert systems are designed to solve complex manufacturing problems by reasoning through bodies of knowledge, represented mainly as if–then rules (K.S.Metaxiotis et al., 2002). For example, Bojan et al. proposed an expert system for finite element mesh generation (B.DOLšAK et al., 1998). The system collected 2000 mesh generation rules to help users select appropriate mesh types and mesh setting parameters. El-Ghany et al. established an expert system for finite element analysis in mechanical manufacturing (K.M.AbdEl-Ghany & M.M.Farag, 2000). The authors introduced human expert experience and established a large number of rules. The system can answer users' questions about geometry, element selection, and boundary condition settings in mechanical manufacturing.

Recently, given the many successful applications of QA system in the open domain, researchers in the manufacturing field have also begun to pay attention to the use of QA systems (Guo & Qing-lin, 2009). For example, Zhang et al. constructed a QA system for automobile manufacturing based on a knowledge graph, and used natural language processing technologies such as entity recognition, entity links, and relationship extraction to improve the accuracy of the QA system (Z.Jingming, 2019). Li et al. built a knowledge-atlas QA system. The QA system can answer the task-based requirements of the machinery industry, assist in analysis, and support decision-making (L.Sizhen, 2019). Chen et al. built a QA system for automobile manufacturing instructions by extracting the semantic structure using a deep neural network, encoding the semantic information in combination with a convolutional neural network (CNN), and using double attention before the pooling layer (C.Conghui, 2020). Li et al. integrated ontology and a mechanical product design scenario into the semantic similarity calculation process, and their QA system can answer users' questions about mechanical product function, structure, principle, index parameters, calculation formula, and engineering material data (L.Hongmei & D.Shengchun, 2015).

# 3 Proposed System

In this section, we present our proposed framework for the technical QA system. As shown in Fig.1, the system consists of two parts: the front end and back end. The front end handles the input of user questions and the output of answers. The back end includes two modules: the knowledge library module and similarity comparison module. The knowledge library module is composed of question-answer pairs. The similarity comparison module finds a set of questions similar to the user's question in the knowledge library and returns the most suitable answer.

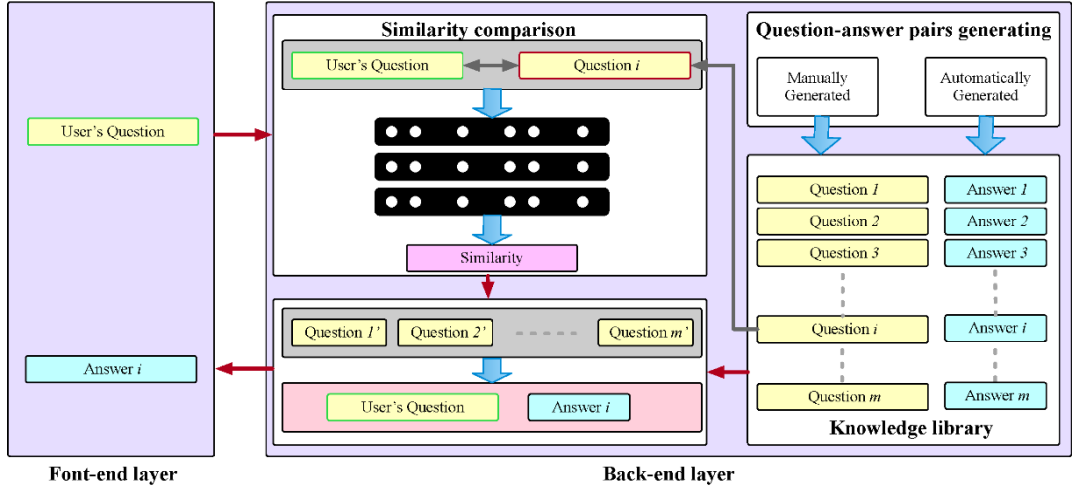

Figure 1: Architecture of the QA system

### 3.1 Question-answer pair generation

The knowledge library module stores question-answer pairs and provides basic data services for the technical QA system. Question-answer pairs are constructed using two methods. One method is to have them generated manually by technical experts according to technical documents.

In this approach, technical experts manually design the basic question-answer pairs $P = \{(q_i, a_i) | i = 1, 2, \cdots, m\}$ according to the standard manual, and then the annotator expands the questions accordingly to obtain $P = \{(q_i^j, a_i) | i = 1, 2, \cdots, m; j = 1, 2, \cdots, k\}$. For example, if the question given by technical experts is "标准适用核能装置的容器吗？(Is the standard applicable to containers for nuclear power plants?)" and the answer is "不能。(No.)," the additional questions created by the annotator include "核能装置的容器适用于本标准吗？(Is the container of nuclear power plant applicable to this standard?)" and "核能装置的容器是否适用于本标准？(Is the container of nuclear power plant applicable to this standard?)[1]," and the answer to these two extended questions is still "不能。(No.)."

The other approach is to automatically generate question-answer pairs using a question word replacement algorithm. We designed a template-based question generation algorithm to automatically generate multiple question-answer pairs from a statement sentence. The algorithm includes the following three steps:

- Carry out named entity recognition (NER) on the standard document, and select the sentences containing entities as candidate statement sentences for generating question-answer pairs.

- Select an appropriate interrogative word according to the entity type, and then replace the entity in the candidate sentence with the interrogative word.

- Adjust the order of words in the sentence containing the interrogative word to form a question. The entity replaced in the previous step will be the answer to the question.

---

[1]The Chinese expressions of these questions differ, but their English translation is the same.

We use the Bert-BiLSTM-CRF model (Y.J.XieT & L.H, 2020) as shown in Fig.2 to realize NER. First, the labeled corpus is transformed into a word vector through BERT pretrained language model. Then input the word vector into the BiLSTM module for further processing. The conditional random field (CRF) module is used to decode the output result of the BiLSTM module to obtain a predictive annotation sequence. Finally, each entity in the sequence is extracted and classified to complete the whole process of NER.

We use the BIO mode to label entities, where B (Begin) represents the starting position of an entity, I (Inside) indicates that the word is inside the entity, and O (Outside) indicates that the word does not belong to any entity. For each entity, we also designed types to describe it. All entity types are listed in Table 1. The Property" indicates the dimension description method of a mechanical product, such as "Geometric dimension." The "Condition" represents the working state of a product in what environment or under what conditions, such as "Corrosive environment." The "Category" refers to the category of the product itself, such as "Evaporator" or "Reactor." The "Material" is the name of the production material of the product. The "Stage" is the stage of the product in the production process, such as "Failure analysis" or "Detection stage."    The "Parameter" refers to a dimension of the product that can be calculated, such as "Length," "Density," or "Roundness."

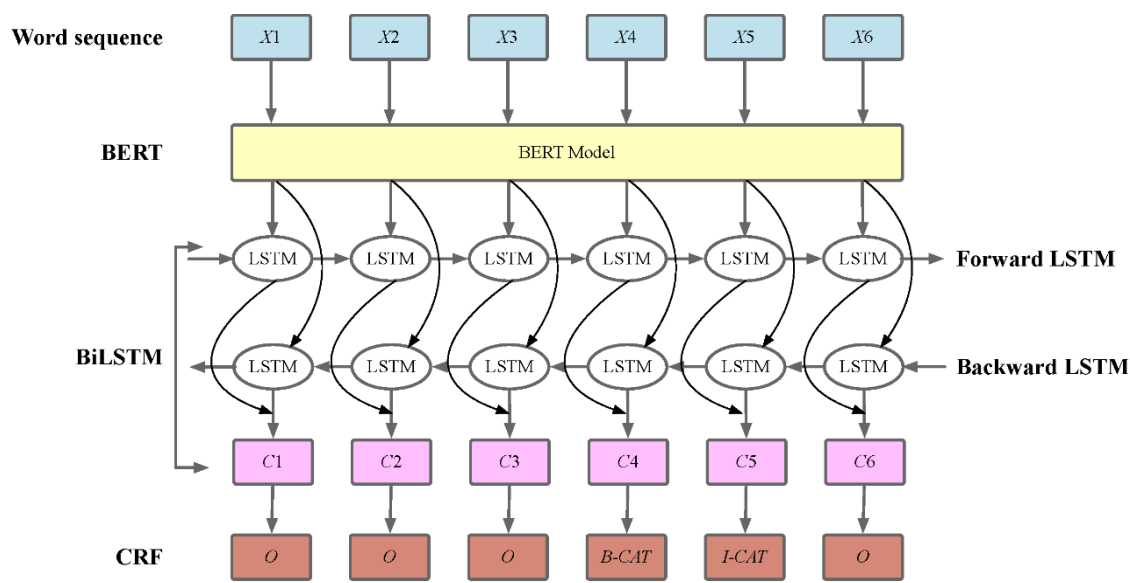

Figure 2: BERT-BiLSTM-CRF network structure

B-PRO indicates that the entity is the starting word of the Property entity. For instance, for the sentence "本规定不适用于对焊法兰的颈部过渡段(This provision is not applicable to the neck transition section of the butt welding flange.)," the entity labels are shown in Table 2.

After determining the entities in the sentence, each entity can be replaced with interrogative words to form a question sentence. Because six types of named entities are used, six interrogative words are used. The six interrogative words correspond to the English terms "what property," "what working condition," "what product category," "what material," "what stage," and "what parameter." The relationship between the named entities and corresponding interrogative words is shown in Table 1.

Table 1: Types of entities.

| Attribute | Abbreviation | Interrogative words |
|---|---|---|
| Property | PRO | 什么属性("what property") |
| Condition | CON | 什么工况("what working condition") |
| Category | CAT | 什么类别("what product category") |
| Material | MAT | 什么材料("what material") |
| Stage | STA | 什么阶段("what stage") |
| Parameter | PAR | 什么参数("what parameter") |

Table 2: Example of label entities.

| sentence | 本 | 规定 | 不 | 适用 | 于 | 对焊 | 法兰 | 的 | 颈部 | 过渡段 |
|---|---|---|---|---|---|---|---|---|---|---|
| label | O | O | O | O | O | B-CAT | I-CAT | O | O | O |

To place the interrogative words in the appropriate position in the question sentence, each word of the question sentence is in a position consistent with Chinese grammar. For the interrogative words "what working condition" and "what parameter" in Table 1, it is necessary to put the interrogative words in the first position of the question sentence. In other cases, according to Chinese grammar, it is possible to directly replace the entity with the interrogative word. For example, for the sentence "本规定不适用于对焊法兰的颈部过渡段(This provision is not applicable to the neck transition section of the butt welding flange.)." The entity of this sentence is "butt welding flange." After replacing the entity with the interrogative word according to Table 1, the question becomes "本规定不适用于什么产品类别的颈部过渡段? (What product category does this provision not apply to the neck transition section?)." After replacement, the formed question-answer pair is ("本规定不适用于什么产品类别的颈部过渡段?", "对焊法兰").

We use PostgreSQL to store question-answer pairs in the knowledge library module. A PostgreSQL database supports JSON format data import. Hence, we can directly import question-answer pairs into PostgreSQL database to facilitate subsequent operations.

### 3.2 Semantic similarity module

For the QA system, when a question $q$ is given, it is necessary to compare $q$ with each question in the question-answer pairs in turn and return the question $s_i$ with the most similar semantics to $q$. The answer $a_i$ of question $s_i$ is used as an alternative answer to question $q$. We proposed an interactive attention BiLSTM (IA-BiLSTM) model, as shown in Fig.3 to compare the similarity between two questions.

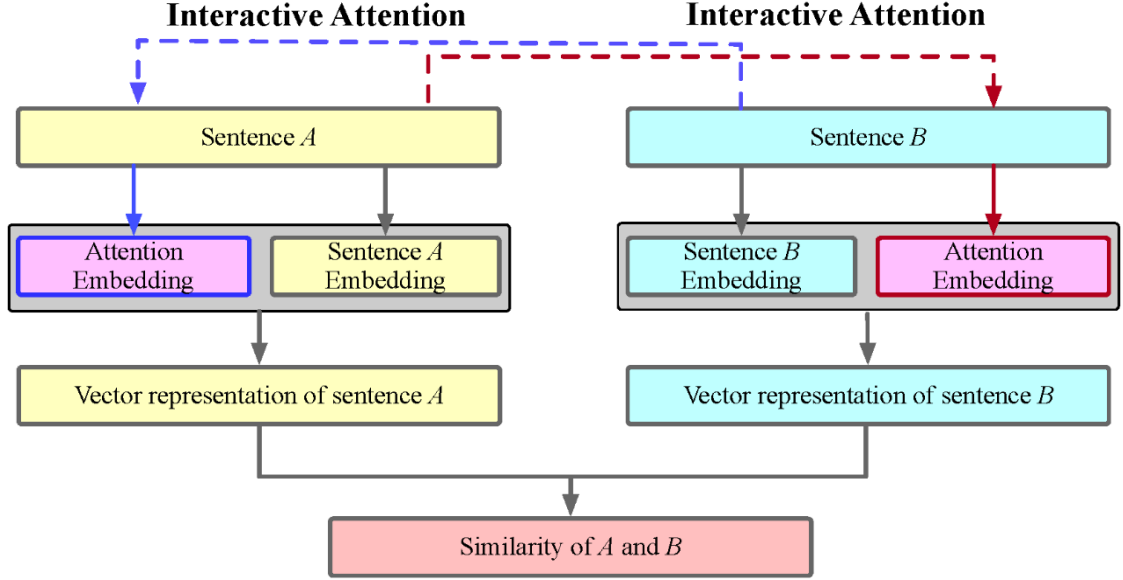

Figure 3: IA-BiLSTM model for computing semantic similarity

The input of this model is two questions $(q_a, q_b)$, and the output is the similarity between them. The similarity value is between 0 and 1, where 0 means that the two sentences are completely different, and 1 means that the two sentences are exactly the same.

The question is segmented using the Jieba word segmentation tool. For example, if the question is "标准适用最大多少压力? (For how much pressure is the standard applicable?)" then the result after word segmentation is "标准/适用/最大/多少/压力." The word vector representation of each word in the sentence can be obtained according to

$$\begin{aligned} a_i &= embedding(q_a^i), \forall i \in [1, \cdots, l_a], \\ b_j &= embedding(q_b^i), \forall i \in [1, \cdots, l_b]. \end{aligned} \quad (1)$$

where *embedding* represents the embedding function, such as word2vec, and $l_a(l_b)$ is the length of words in sentence $q_a(q_b)$. We then use BiLSTM to encode the word context in the sentence. A BiLSTM runs a forward and backward LSTM on a sequence starting from the left and right ends, respectively. The hidden states generated by these two LSTMs at each word are concatenated to represent a word and its context. We write $\bar{a}_i(\bar{b}_j)$ to denote the output state generated by the BiLSTM at $i$th ($j$th) word over the input sequence $q_a(q_b)$.

$$\begin{aligned} \bar{a}_i &= \text{BiLSTM}(a, i), \\ \bar{b}_j &= \text{BiLSTM}(b, i). \end{aligned} \quad (2)$$

After the word vector containing the context is obtained, the cross attention can be calculated according to the word vector. We calculate the dot product between two word vectors to represent the attention weight between them, that is,

$$e_{ij} = \bar{a}_i^T \times \bar{b}_j. \tag{3}$$

For the $i$th word in sentence $q_a$, we accumulate each word vector in sentence $q_b$ to form a new representation of word $i$, which is represented as $\tilde{a}_i$ with Eq.4. The word vector in sentence $q_b$ is multiplied by the attention weight in Eq.3 so that the word in sentence $q_b$ is most related to the $i$th word in sentence $q_a$. The same is applied to $\tilde{b}_j$ with Eq.4.

$$\begin{aligned}\tilde{a}_i &= \sum_{j=1}^{l_b} \frac{\exp e_{ij}}{\sum_{k=1}^{l_b} \exp e_{ik}} \bar{b}_j, \\ \tilde{b}_j &= \sum_{i=1}^{l_b} \frac{\exp e_{ij}}{\sum_{k=1}^{l_a} \exp e_{kj}} \bar{a}_i.\end{aligned} \tag{4}$$

The sentence vector generated by BiLSTM and the sentence vector obtained by interactive attention are combined to form the final vector representation of the sentence, that is,

$$\begin{aligned}v_a &= [\bar{a}, \tilde{a}], \\ v_b &= [\bar{b}, \tilde{b}].\end{aligned} \tag{5}$$

The similarity of two sentences is calculated by cosine similarity. The cosine similarity formula of vector $v_a$ and vector $v_b$ is as follows:

$$d_{ab} = \frac{\sum_{i=1}^{n} v_a^i v_b^i}{\sqrt{\sum_{i=1}^{n} (v_a^i)^2} \sqrt{\sum_{i=1}^{n} (v_b^i)^2}}. \tag{6}$$

## 4 Results

### 4.1 Performance of question-answer pair generation

We first evaluate the performance of the question-answer pair generating algorithm. The proposed system has 3,649 question-answer pairs that were automatically generated according to our method and 7,601 manually generated question-answer pairs (see Section 3.1). Technical experts and annotators were recruited to automatically generate question-answer pairs. In this study, the technical experts were master students majoring in mechanical engineering, and the annotators were undergraduates majoring in mechanical engineering. All students were recruited from the Zhejiang University of Technology.

Table 3: Scoring rules for the four indicators.

| Score | Relevance | Fluency | Ambiguity | Instruction |
|---|---|---|---|---|
| 1 | Completely irrelevant. | Total Inconsistent. | Contradictions. | No help. |
| 2 | Partial correlation. | Partial Inconsistent. | Partial Contradictions. | Partial help. |
| 3 | Most related. | Most Consistent. | One ambiguous word. | Helps. |
| 4 | Completely related. | Fluency. | No ambiguous word. | Helps a lot. |

We manually evaluated these question-answer pairs using the four indicators of relevance, fluency, ambiguity, and instruction. The scoring rules for these four indicators are listed in Table 3. We recruited a total of 30 volunteers majoring in mechanical engineering to score the question-answer pairs, and the average scores are shown in Table 4. Compared with the manually generated question-answer pairs, automatically generated question-answer pairs have better relevance and instruction, but score less well with respect to fluency and ambiguity. In addition, the method of automatically generating question-answer pairs does not consider the relevance of context when replacing question words, resulting in a low score for the ambiguity indicator. Although the indicator results of automatically generated question-answer pairs are generally lower than those of manually generated question-answer pairs, considering the high cost of manually generated question-answer pairs, automatically generated question-answer pairs can balance the trade-off between the cost and size of a training dataset.

Table 4: Average score of generated questions.

|  | Relevance | Fluency | Ambiguity | Instruction |
| --- | --- | --- | --- | --- |
| Manually generated | 3 | 3.3 | 3.4 | 2.8 |
| Automatically generated | 3.5 | 2.9 | 2.3 | 3.4 |

### 4.2 Performance of semantic similarity between questions

We compared the performance of the IA-BiLSTM model and existing models such as CNN, LSTM, and BiLSTM on the problem semantic similarity matching task using the Quora question pairs (Qqp) dataset. The Qqp dataset is a collection of similar question pairs found in Quora. The question pairs in the training set are marked as similar or dissimilar, and there are 363,870 pairs in the training set, 390,965 pairs in the test set, and 40,431 pairs in the verification set. We used the CNN, LSTM, and BiLSTM models for comparison with the proposed IA-BiLSTM model. Each model was trained for 20 iterations. After each training process, the test dataset was used to test the accuracy of the model.

Alibaba cloud's Platform of Artificial Intelligence (PAI) was used to run these experiments, and the neural network model training code was written in the Notebook Interactive Data Science Workshop. The hardware environment was the ecs.gn5-c4g1.xlarge instance on the PAI platform, which contains four virtual CPUs, 30G of memory, a NVIDIA P100 (graphics card), and 3 Gbps bandwidth. The experimental data were divided into training and test sets according to the ratio of 9:1. Nesterov adaptive motion estimation was used as the optimization function to adjust the weights of the neural networks, and cross entropy was used as the loss function. The TensorFlow library was used to build the neural network models, scikit-learn was used for the dataset segmentation, and the Jieba word splitter was used for sentence segmentation

As shown in Table 5, the accuracies of the LSTM, BiLSTM, and CNN models under this dataset are approximately 72%, which is lower than that of IA-BiLSTM. The accuracy of the BiLSTM model with attention is 73.01%, whereas the IA-BiLSTM model proposed in this paper has the highest accuracy (74.42%).

Table 5: Performance comparison with other models.

| Model | Accuracy |
| --- | --- |
| CNN | 0.7291 |
| LSTM | 0.7238 |
| BiLSTM | 0.7223 |
| **IA-BiLSTM** | **0.7442** |

## 4.3 Case study

According to the method proposed above, we developed a QA system based on the "JB4732 steel pressure vessels - analysis and design standard" document. JB4732 is a professional mandatory standard for pressure vessels, which is reviewed and approved by the National Technical Committee of China for pressure vessel standardization. The applicable scope of the standard includes vessels with a design pressure greater than or equal to 0.01 MPa and less than 100 MPa and vessels with a vacuum degree greater than or equal to 0.02 MPa. The standard considers the design, material selection, manufacturing, inspection, and acceptance of vessels as a system, and gives the methods of stress analysis and fatigue analysis. The standard has 11 chapters and 11 appendices.

We designed a QA system based on the JB4732 standard to help engineers quickly determine various common problems in the design process. Considering the modularity of the system implementation, we further divided the functional structure shown in Fig.1 into functional modules. The front end of the system includes an interaction module, and the back end of the system contains a function module, model module, and data module. The interaction module is mainly responsible for processing the input of administrators and users. The function module provides various services for the interaction module, such as log viewing, model training, user response, and QA knowledge library update. These services are implemented in the function module and encapsulated as independent functions for retrieval by the interaction module. The model module mainly provides technical support for the function module, including data preprocessing, NER, and similarity matching sub-modules. The data module provides data storage services for the model module, including the standard question table, historical user question table, feedback table, and entity table.

The user interface of the system is shown in Fig.4. After the user enters the question in the search box, the system will give the corresponding candidate answer and the location of the answer in JB4732 standard. The user clicks the "jump" button to go the page in the JB4732 standard where the answer is located to help the user verify the answer.

We used the system to test 1,785 problems, which were categorized into four groups (design method, concept definition, numerical value, and calculation method), and the accuracy rate of the results was 98.15%.

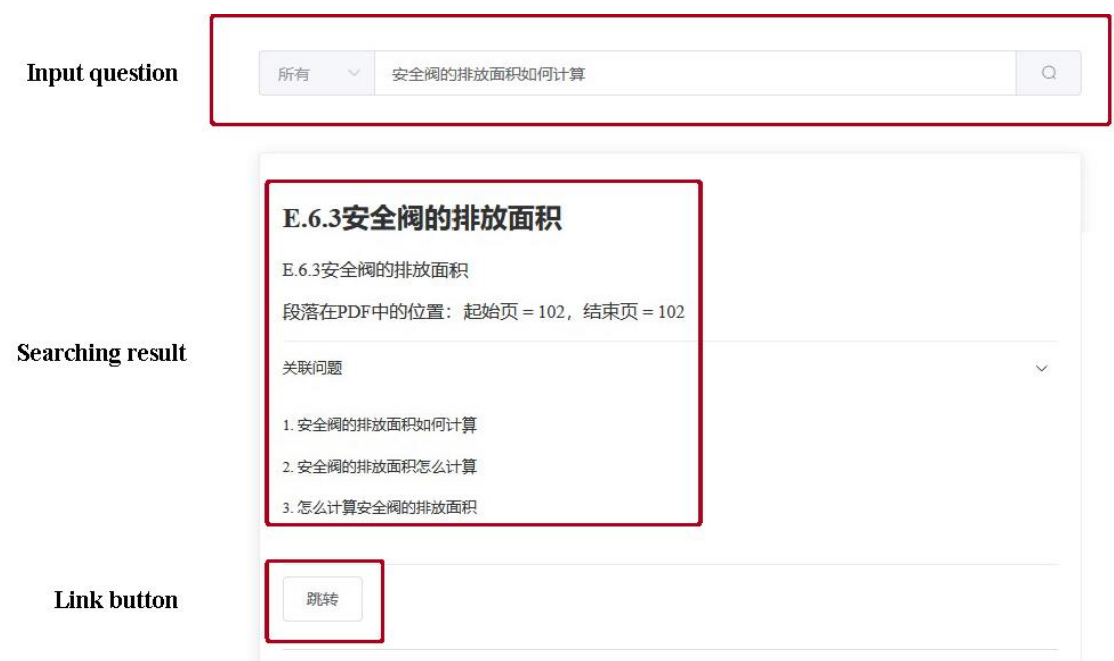

Figure 4: Interface of the QA system

For example, when the user enters a question belonging to the design method: "安全阀的排放面积如何计算？(How is the discharge area of the safety valve calculated? )," the similar question returned by the system is "怎么计算安全阀的排放面积？(How is the discharge area of the safety valve calculated?) ." The system then outputs the location Chapter E.6.3 and a link button. Users click on the button to open the page that includes the method of calculating the discharge area. When the user enters a numerical question such as "制造单位的容器技术文件至少保存几年？(How many years must the container technical documents of the manufacturing unit be kept?)," the system directly returns the answer to this question: 5 years.

# 5 Conclusions

The design and manufacturing process of special mechanical products must comply with the national standards, and engineers must repeatedly query the standard in the design process. At present, the knowledge retrieval methods based on expert systems and keyword queries are not only expensive to maintain, but also cannot automatically give targeted answers to problems. With the development of natural language techniques, the integration of intelligence in knowledge queries in the field of special machinery is becoming more common. To improve the performance of a QA system, it is necessary to automatically construct a dataset for model training and to improve the performance of sentence similarity comparison. To address these two tasks, this study proposed a method that can automatically construct question-answer pairs. In addition, a IA-BiLSTM model integrating BiLSTM and interactive attention was designed to calculate the semantic similarity between two sentences. Finally, based on the proposed method, we developed a QA system for the JB4732 standard. The results show that the system can enable pressure vessel designers to quickly determine the answer to common problems in the design process. The proposed methods in this paper could be applied as useful supporting tools for the construction of automatic QA systems in the mechanical manufacturing field.

Although the methods proposed in this study have substantially improved the performance of a technical QA system, there are still some limitations; new questions cannot be answered directly according to the contents of the standard. Therefore, in future research, we plan to integrate document understanding and reasoning technologies into the technical QA system to improve the flexibility of the system.

# Acknowledgements

This work is supported by the National Science Foundation of China (No. 61976193) and the ZheJiang Natural Science Foundation of China (No. LY19F020034). We thank Kimberly Moravec, PhD, from Liwen Bianji (Edanz) (www.liwenbianji.cn/) for editing the English text of a draft of this manuscript.